# From Manual Observation to Automated Monitoring: Space Allowance Effects on Play Behaviour in Group-Housed Dairy Calves

Authors: H. Yang[1], H. Lesscher[2], Liu[1], M. Hostens[1]

Institution:

[1] Cornell University, Department of Animal Science, College of Agriculture and Life Sciences, Ithaca, NY 14853

[2] Utrecht University, Department Population Health Sciences, Veterinary Medicine, 3584 CM Utrecht

# Abstract

Play behaviour serves as a positive welfare indicator in dairy calves, yet the influence of space allowance under commercial conditions remains poorly characterized, particularly at intermediate-to-high allowances (6-20 $m^2$ per calf). This study investigated the relationship between space allowance and play behaviour in 60 group-housed dairy calves across 14 commercial farms in the Netherlands (space range: 2.66-17.98 $m^2$ per calf), and developed an automated computer vision pipeline for scalable monitoring. Video observations were analyzed using a detailed ethogram, with play expressed as percentage of observation period (%OP). Statistical analysis employed linear mixed models with farm as a random effect. A computer vision pipeline was trained on manual annotations from 108 hours on 6 farms and validated on held-out test data. The computer vision classifier achieved 97.6% accuracy with 99.4% recall for active play detection. Calves spent on average 1.0% of OP playing reflecting around 10 minutes per 17-hour period. The space-play relationship was non-linear, with highest play levels at 8-10 m$^2$ per calf (1.6% OP) and lowest at 6-8 m$^2$ and 12-14 m$^2$ (<0.6% OP). Space remained significant after controlling for age, health, and group size. In summary, these findings suggest that 8-10 m$^2$ per calf represents a practical target balancing welfare benefits with economic feasibility, and demonstrate that automated monitoring can scale small annotation projects to continuous welfare assessment systems.

# 1. Introduction

Animal welfare has become central to modern dairy production, driven by ethical responsibility, consumer expectations, and sustainability considerations (Miller-Cushon & Jensen, 2025). Scientific focus has shifted from preventing negative outcomes to creating conditions enabling positive emotional states and species-typical behaviours (Mellor, 2016). Young calves are especially sensitive to environmental conditions during critical developmental periods, with early-life experiences shaping long-term health, productivity, and behavioural competence (Costa et al., 2016). Current European Union legislation mandates minimum space allowances of 1.5-1.8 m$^2$ per calf (Council Directive 2008/119/EC, 2008), but the European Food Safety Authority recently recommended 20 m$^2$ per calf for full locomotor play expression (EFSA AHAW Panel, 2023). This discrepancy, combined with limited empirical data at intermediate space allowances under commercial conditions, motivates investigation of the space-play relationship across a broader gradient.

Play behaviour, i.e. voluntary and repetitive activity serving no immediate survival function and performed in relaxed contexts, has gained recognition as a valuable positive welfare indicator in dairy calves (Held & Špinka, 2011). Play is manifested by calves as locomotor (running, jumping, kicking), social (frontal pushing, mounting), and object play (interacting with pen features), supporting neuromuscular development, social learning, and cognitive growth (Jensen et al., 1998). Play is suppressed by stressors including disease, hunger, and fear (Mintline et al., 2013), making it a potential sensor for compromised welfare. Multiple studies demonstrate that larger pens promote more play than smaller pens

(Jensen et al., 1998; Jensen & Kyhn, 2000), attributed to reduced physical constraints, enhanced social facilitation, and improved affective states. However, most research examined 1-6 m$^2$ per calf in experimental settings, leaving substantial gaps regarding intermediate-to-high allowances (6-20 m$^2$) under commercial conditions where multiple factors interact.

Despite the recognized value of play behaviour, systematic monitoring on commercial farms remains rare due to methodological limitations. Manual observation is labor-intensive, subject to observer bias, and lacks scalability (Martin & Bateson, 2007). Accelerometer-based approaches offer continuous monitoring but exhibit low specificity for play, confounding play-related movement with feeding and other non-play activities. Gladden et al. (2020) found a correlation with manual observations of 0.48 to 0.70. Accelerometers also fail to distinguish play types and provide no spatial context for interpreting welfare implications.

Computer vision offers a promising alternative, leveraging cameras and deep learning algorithms to automate behaviour recognition. Recent foundation models pre-trained on massive datasets using self-supervised learning have dramatically improved accuracy and domain transfer. For example, YOLOv12 enables real-time object detection, SAM2 provides promptable segmentation and tracking in video, and DinoV2 generates rich visual embeddings transferable to downstream tasks with minimal task-specific training (Oquab et al., 2024; Ravi et al., 2024). These models have been applied to livestock monitoring for feeding, rumination, and activity detection in adult cattle (Bezen et al., 2020), and more recently to play behaviour detection in pigs with >95% accuracy (Yang et al., 2025). However, application to dairy calves—smaller, faster-moving, with higher social interaction rates—remains limited, and validation under commercial farm conditions (with variable lighting, diverse pen layouts, occlusions) is scarce.

This study pursued three objectives: (1) quantify the relationship between space allowanceand play behaviour under commercial conditions using detailed manual annotation and linear mixed models; (2) develop and validate an automated computer vision pipeline integrating state-of-the-art foundation models, achieving >90% accuracy for practical welfare monitoring; and (3) provide evidence-based recommendations for implementation. By integrating rigorous behavioural observation with technological innovation, we aimed to advance both animal welfare science and practical farm management, demonstrating that automated systems can scale small annotation projects to continuous, objective welfare assessment.

# 2. Materials and Methods

## 2.1 Study Design and Farm Selection

This study investigated the relationship between space allowance and play behaviour in group-housed dairy calves across commercial farms in the Netherlands. Data collection took place between June 2024 and May 2025, covering 14 farms with 20 pens (one farm excluded due to incomplete video processing). Each farm visit lasted approximately one week with continuous video recording of all calves in selected pens.

Farms were selected to provide variation in space allowance (2.66-17.98 $m^2$ per calf) while maintaining comparable housing systems (group housing with straw bedding). Selection criteria included: (1) pen sizes ranging from <4 to >16 $m^2$ per calf, (2) group configurations with at least 2 calves per pen, (3) farmer consent for camera installation and continuous recording, and (4) pen layouts allowing adequate camera coverage. The study was conducted in accordance with the Directive 2010/63/EU. Ethical approval was obtained from the animal welfare body at Utrecht University, The Netherlands.

## 2.2 Housing Conditions and Animal Characteristics

### 2.2.1 Pen Characteristics

Pen dimensions were measured manually during each visit. Total surface area was calculated with inaccessible areas (e.g., milk feeder bases) subtracted. Space allowance per calf was calculated by dividing usable floor space by the number of calves. Total pen areas ranged from 10.63 to 70.33 $m^2$ (mean = 32.06 $m^2$, SD = 16.79 $m^2$), yielding space allowances of 2.66 to 17.98 $m^2$ per calf (mean = 9.58 $m^2$, SD = 4.50 $m^2$). For analysis, the data was categorized based on 2 $m^2$ increments in space : <4, 4-6, 6-8, 8-10, 10-12, 12-14, and 16-18 $m^2$ per calf. No calves could be included with 14-16 $m^2$.

All pens featured straw bedding on hard flooring (93.3%) or mixed hard-rubber flooring (6.7%). Bedding quality was assessed twice per visit and scored as: 1 (clean both times), 2 (clean once, moderately dirty once), or 3 (moderately dirty both times).

## 2.2.2 Animal Characteristics

The final sample included 64 calves that were predominantly female (98.3%) and Holstein Friesian (81.7%), with 18.3% crossbreeds. Age ranged from 2 to 114 days (mean = 52.6, SD = 25.5), categorized as: under 1 month (21.7%), 1-2 months (41.7%), 2-3 months (21.7%), and over 3 months (15.0%). Estimated body weights ranged from 42 to 157 kg (mean = 88.5 kg, SD = 28.3 kg). Average daily milk intake ranged from 0 to 18 L (mean = 5.89 L, SD = 3.60 L), reflecting age-related differences in nutritional needs.

*Table 1. Descriptive characteristics of study population (n=64 calves across 14 farms). Summary statistics for calf demographics, housing conditions, and play behaviour observations.*

| Variable | Minimum | Maximum | Mean | SD |
|---|---|---|---|---|
| **Calf characteristics** | | | | |
| Age (days) | 2.00 | 114.00 | 53.28 | 25.88 |
| Body weight (kg) | 47.00 | 166.00 | 87.36 | 27.76 |
| Milk intake (L/day) | 0.00 | 18.00 | 5.59 | 3.57 |
| **Housing characteristics** | | | | |
| Pen size (m$^2$) | 10.63 | 70.33 | 30.68 | 16.65 |
| Space per calf (m$^2$) | 2.66 | 17.57 | 8.98 | 4.27 |
| Group size (calves/pen) | 2.00 | 6.00 | 3.59 | 1.22 |

## 2.2.3 Health Status Assessment

Health status was systematically assessed at each visit. Calves with severe illness (Category 4) were excluded to avoid handling stress and because severe illness suppresses play behaviour (Ahloy-Dallaire et al., 2018). Calves vaccinated or dehorned within two weeks were also excluded due to temporary behavioural alterations (Mintline et al., 2013). Enrolled calves were classified as: Category 1 (Healthy, n=37, 57.8%) or Category 2 (Mildly unhealthy with single minor symptoms, n=21, 32.8%) or Category 3 calves (multiple concurrent symptoms, n=6, 9.4%).

# 2.3 Video Data Collection

## 2.3.1 Recording Equipment and Setup

Calf behaviour was recorded continuously using Bascom camera systems (Bascom Cameras B.V., Nieuwegein, Netherlands). Cameras were installed to provide full pen visibility, mounted on ceilings, poles, or structural supports. Recording began 06:00 and ended 23:00 daily.

## 2.3.2 Video Storage and Processing

Videos were stored on SD cards (128 GB or 256 GB capacity) inserted into each camera unit. At visit conclusion, SD cards were collected and video files were transferred to external hard drives. Video files were organized by farm, pen, camera, and date. Further technical details on video formats and processing workflows are provided in Supplementary Material S1.

## 2.4 Manual Annotation (Ground Truth Creation)

### 2.4.1 Annotation Protocol

All video footage was reviewed manually by 7 trained observers using Observer XT software (Noldus Information Technology, 2023). Observers were trained with a standardized ethogram to ensure consistent behaviour classification. For each farm, one complete 17-hour observation period (06:00-23:00) was analyzed, focusing on the second day to minimize researcher presence effects. Videos were reviewed at variable speeds: normal or reduced speed (0.5×-1×) when play occurred, and increased speed (2×-4×) during inactive periods. External disturbances (e.g., farm personnel entering pens) were documented with timestamps.

### 2.4.2 Ethogram Development

The ethogram focused on play behaviours categorized into four subtypes: locomotor play (running, jumping, kicking, bucking), social play (frontal pushing, mounting, chasing), object play (interacting with pen features), and straw-related play. Definitions were adapted from Jensen et al. (1998) and expanded to include additional behaviours observed in commercial settings. Non-play states ("management," "out of view") were adapted from Gladden et al. (2020), with farm-specific categories ("individual," "milk feeding") added. The complete ethogram with operational definitions is provided in Supplementary Table S1.

### 2.4.3 Ground Truth Format

Behavioural observations were recorded with 0.1-second temporal resolution. Observer XT output was exported as CSV files containing: Subject (calf ID), Behaviour (play type or non-play state), Modifier (behavioural variants), Event_Type (state start/stop), Time_Relative_sf (seconds from video start), and Duration (seconds). This format enabled precise temporal alignment with video frames for computer vision training.

## 2.5 Statistical Analysis

### 2.5.1 Play Behaviour Quantification

Play duration was calculated by summing all play bout durations for each calf over the 17-hour observation period. Total play time was expressed as percentage of observation period (%OP), calculated as: (total play seconds / total observation seconds) × 100. This metric normalized play across calves with slight variations in observation timing.

### 2.5.2 Statistical Models

Statistical analyses were performed using SPSS (v29, IBM Corp.). Linear mixed models examined the relationship between space allowance and play behaviour, with farm as a random effect to account for clustering. The base model tested space category as the sole fixed effect. Model diagnostics verified assumptions of normality and homoscedasticity. A multivariable model subsequently examined space effects after controlling for age, health status, milk intake, group size, and bedding quality. Model selection employed likelihood ratio tests, AIC, and BIC. Statistical significance was assessed at alpha=0.05. Detailed model specifications, equations, and diagnostic procedures are provided in Supplementary Material S3.

## 2.6 Computer Vision Pipeline Development

### 2.6.1 Pipeline Overview

The computer vision system integrated three state-of-the-art foundation models: YOLOv12 for calf detection (Tian et al., 2025), SAM2 for segmentation and tracking (Ravi et al., 2024), and DinoV2 for feature extraction (Oquab et al., 2023). This pipeline automated play behaviour detection from continuous video footage, enabling scalable welfare monitoring beyond manual annotation capacity.

## 2.6.2 Object Detection and Tracking

Video frames were extracted at 1 fps and processed through YOLOv12 (pre-trained on COCO dataset) to detect calf bounding boxes with confidence ≥0.55. SAM2 generated segmentation masks for each detected calf and tracked individuals across frames using video memory and object persistence mechanisms. Each tracked calf received a unique ID maintained throughout sequences. Tracking was reset when calves remained out of view >5 seconds to prevent ID reassignment errors.

## 2.6.3 Feature Extraction

Cropped calf images (based on segmentation masks) were processed through DinoV2 (vision transformer pre-trained via self-supervised learning on 142 million images) to generate 1024-dimensional embedding vectors. These embeddings captured rich visual features without task-specific training, providing robust representations for downstream classification.

## 2.6.4 Metadata Integration

The final training dataset required integrating three data streams: (1) manual behavioral annotations from Observer XT, (2) video frames with detected calves and tracking IDs, and (3) DinoV2 embeddings. This integration involved several steps executed using Apache Spark for distributed processing of large-scale data.

**Timestamp extraction:** Video timestamps were extracted via optical character recognition (OCR) from on-screen displays burned into video frames. OCR outputs were validated against video metadata and corrected for parsing errors to ensure temporal accuracy.

**Temporal alignment:** Manual annotations (recorded in seconds from video start) were converted to absolute timestamps and matched to extracted frames within a 0.5-second tolerance window. This tolerance accommodated minor discrepancies between annotation precision and frame extraction timing.

Output format: The final metadata file linked each frame to its corresponding behavioral label, tracking ID, cropped image path, and embedding path, creating training samples that connected visual features to behavior classifications. Detailed metadata construction logic, data engineering procedures, and quality assurance steps are provided in Supplementary Material S4.

### 2.6.5 Frame Exclusion Criteria

Not all recorded footage was suitable for training. Exclusions included: (1) sequences where calves were out of view >5 seconds (tracking failures), (2) frames with severe occlusion by pen structures (>50% of bounding box), (3) insufficient resolution (bounding box <100 pixels in smallest dimension), and (4) extreme lighting conditions (mean pixel intensity <30 or >225). These criteria yielded 108 hours of usable footage from 6 farms for pipeline training, with approximately 1.5 million frame-label pairs. Further details on exclusion impacts are provided in Supplementary Material S5.

## 2.7 Classification Model Training

### 2.7.1 Dataset Preparation

The initial dataset exhibited class imbalance reflecting natural behaviour prevalence ("Not Playing" majority, "Active Playing" minority). To prevent model bias toward the majority class, we implemented stratified downsampling to create a balanced dataset with equal representation of three classes: Active Playing, Non-Active Playing, and Not Playing. The balanced dataset (22,827 samples total, 7,609 per class) was split into training (70%), validation (15%), and test (15%) sets using stratified sampling to maintain class balance across splits.

### 2.7.2 Model Architecture and Training

A multi-layer perceptron (MLP) classifier was trained on DinoV2 embeddings to predict behaviour class. The architecture comprised: input layer (1024 dimensions), two hidden layers (512 and 256 units with ReLU activation and 50% dropout), and output layer (3 classes). The model was trained using Adam optimizer (learning rate=0.001, weight decay=0.00001) with cross-entropy loss and early stopping (patience=5 epochs). Training was monitored using validation loss, with the best checkpoint selected for final evaluation. Detailed architectural specifications, hyperparameters, and training procedures are provided in Supplementary Material S6 and the GitHub repository.

### 2.7.3 Model Evaluation

The best model was evaluated on the held-out test set. Performance metrics included overall accuracy, per-class precision, recall, and F1-score, plus confusion matrices showing classification patterns. All experiments were tracked using MLflow for reproducibility (Zaharia et al., 2018).

## 2.8 Computing Environment and Software

All computational analyses were performed on dedicated computing infrastructure. Statistical analyses were conducted using IBM SPSS Statistics (Version 29, IBM Corp., Armonk, NY, USA). Computer vision pipeline development and model training utilized Python 3.10 with the following key libraries: PyTorch 2.0 for deep learning, OpenCV 4.8 for image processing, Apache Spark 3.4 for distributed data processing, and scikit-learn 1.3 for evaluation metrics.

Video frame extraction and preprocessing were performed on a Databricks cluster equipped with an Intel Xeon processor and 64 GB RAM. Model training utilized NVIDIA V100 GPUs (32 GB VRAM) on a Databricks cluster (Standard_NC6s_v3 instances, 1 GPU, 112 GB RAM per node, 32 GB VRAM, 6 cores, NVIDIA Tesla V100 GPUs). ll code for the computer vision pipeline and analysis scripts is available on the [GitHub repository](https://github.com/Bovi-analytics/Individual-Behavior-Analysis-with-CV): [https://github.com/Bovi-analytics/Individual-Behavior-Analysis-with-CV].

# 3. Results

## 3.1 Farm and Animal Characteristics

**Space allowance distribution.** Space per calf varied substantially: 2.66-17.57 $m^2$ (mean = 8.97 $m^2$, SD = 4.27 $m^2$). The distribution across 2 $m^2$ increment categories was: <4 $m^2$ (n=9, 11.7%), 4-6 $m^2$ (n=10, 13.3%), 6-8 $m^2$ (n=9, 13.3%), 8-10 $m^2$ (n=10, 16.7%), 10-12 $m^2$ (n=11, 18.3%), 12-14 $m^2$ (n=8, 11.7%), and 16-18 $m^2$ (n=7, 15.0%). No calves were housed at 14-16 $m^2$ per calf.

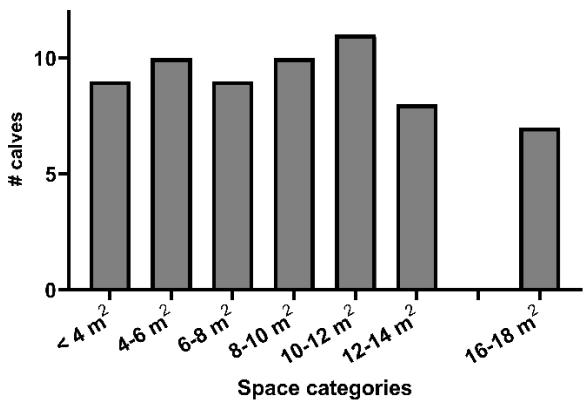

*Figure 1. Distribution of calves (n = 64) across space allowance categories measured in m² per calf. Space allowances were categorized in 2 m² increments from <4 m² to 16-18 m² per calf. Each bar represents the number of calves housed at the corresponding space allowance.*

**Demographic characteristics.** The sample was predominantly female (96.9%) and Holstein Friesian (85.9%), with 14.1% crossbreeds. Age ranged from 2 to 114 days (mean = 53.3, SD = 25.9), distributed as: <1 month (23.4%), 1-2 months (39.1%), 2-3 months (28.1%), and >3 months (9.4%). Body weight averaged 87.4 kg (SD = 27.8 kg, range 47-166 kg). Daily milk intake ranged from 0 to 18 L (mean = 5.6 L, SD = 3.6 L), reflecting developmental stage differences. Health status was classified as healthy (n=37, 57.8%), mildly unhealthy (n=21, 32.8%) or moderately unhealthy (n = 6, 9.4%).

**Housing characteristics.** All pens featured straw bedding on hard flooring (92.2%) or mixed hard-rubber flooring (7.8%). Bedding quality was clean twice (both at the onset and end of the observation week, 37.5%), clean once and moderately dirty once (either at the onset or end of the observation week, 37.5%), or moderately dirty twice (25.0%).

# 3.2 Play Behaviour Descriptive Statistics

## 3.2.1 Overall Play Patterns

Calves spent on average 1.0% of the 17-hour observation period engaged in play (SD = 0.68%, range 0.08-3.13%), corresponding to approximately 10 minutes per day. Individual variation was substantial: the most playful calf played 31.9 minutes (3.13% OP), while the least playful played only 49 seconds (0.08% OP), representing a 39-fold range.

**Play composition by type.** Social play constituted the largest component (mean = 0.55% OP, 55% of total play), followed by locomotor play (0.26% OP, 26%), object play (0.17% OP, 15%), and straw play (0.02% OP, 2%). This distribution indicates calves preferentially engaged in social interactions during play.

**Frequency of play events.** Calves engaged in 6.70 play events per hour on average (SD = 5.38, range 0.78-32.17). Locomotor play occurred most frequently (4.31 events/hour), followed by social play (1.82 events/hour), object play (0.46 events/hour), and straw play (0.06 events/hour). The high frequency of locomotor play relative to its moderate duration suggests these behaviours were executed in brief episodes (e.g., single 2-3 second gallops), whereas social play events tended to last longer.

### 3.2.2 Specific Behaviour Frequencies

Frequencies of specific play behaviours are summarized in Table 3. Among locomotor behaviours, galloping was most common (1.30 events/hour, SD = 1.80, maximum 12.80), followed by leaping (0.78 events/hour) and leaping sideways (0.65 events/hour). Less frequent were turning (0.12 events/hour), jumping (0.17 events/hour), bucking (0.29 events/hour), and kicking (0.19 events/hour). For social play, frontal pushing dominated (1.69 events/hour, SD = 1.46, maximum 6.95), while mounting was rare (0.13 events/hour). Object play behaviours included butting fixtures (0.33 events/hour) and head-butting movable objects (0.13 events/hour). Head-shaking classified as play averaged 0.82 events per hour.

**Table 3. Frequency of individual play behaviours (events per hour, n=64 calves).**

| Behaviour Type | Behaviour | Mean ± SD | Range |
| --- | --- | --- | --- |
| **Locomotor Play** | Gallop | 1.30 ± 1.80 | 0-12.80 |
| | Leap | 0.78 ± 1.03 | 0-7.20 |
| | Leap sideways | 0.65 ± 0.83 | 0-5.12 |
| | Buck | 0.29 ± 0.41 | 0-2.60 |

|  | Kick | 0.19 ± 0.24 | 0-1.14 |
|  | Jump | 0.17 ± 0.25 | 0-1.28 |
|  | Turn | 0.12 ± 0.22 | 0-1.07 |
| **Social Play** | Frontal push | 1.69 ± 1.46 | 0-6.95 |
|  | Mounting | 0.13 ± 0.51 | 0-3.90 |
| **Object Play** | Butt fixtures | 0.33 ± 0.46 | 0-2.65 |
|  | Head-butt objects | 0.13 ± 0.28 | 0-1.76 |
| **Other** | Head-shake (play) | 0.82 ± 0.74 | 0-3.88 |
|  | Straw interaction | 0.06 ± 0.14 | 0-0.79 |

## 3.3 Space Allowance Effects on Play Behaviour

### 3.3.1 Univariate Analysis: Space and Total Play

Space allowance significantly affected total play duration ($P < 0.01$, Figure 2). Fixed effects (space category alone) accounted for 20,8% of the variance. The relationship was non-linear: lowest play occurred at 6-8 m² (0.60% OP, ~6 minutes) and 12-14 m² (0.40% OP, ~4 minutes), while highest play occurred at 8-10 m² (1.59% OP, ~16 minutes). Intermediate-to-high play was also observed at 10-12 m² (1.35% OP) and 16-18 m² (1.29% OP).

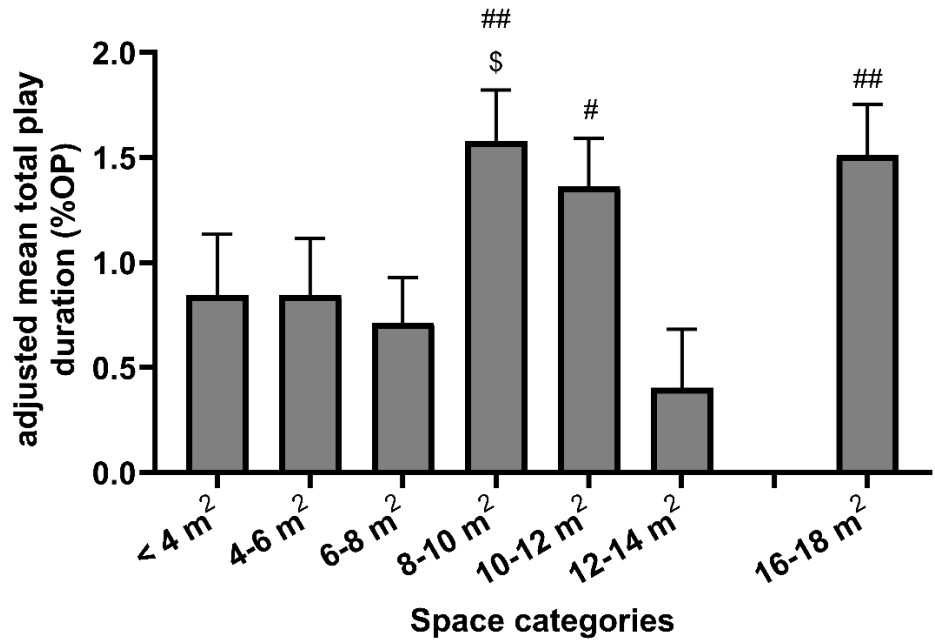

*Figure 2. Model-adjusted mean total play duration (expressed as percentage of the observation period, %OP) for dairy calves across seven space allowance categories. Bars represent means + SEM from a linear mixed model with farm as a random effect (n = 64 calves). Play duration was highest in the 8-10 m², 10-12 m² and 16-18 m² categories . Significance by LSD pairwise comparisons are indicated by $ (P < 0.05) for comparisons with calves in the 4-6 m² category and # or ## (P < 0.05 or P < 0.01) for comparisons with calves in the 6-8 m² category.*

## 3.3.2 Space and Locomotor, Social and Object Play

Contrary to expectations, space allowance was not a predictor of locomotor play duration (Figure 3A). Model-adjusted means ranged from 0.11% OP (<4 m²) to 0.54% OP (10-12 m²), with calves in the 10-12 m² and 16-18 m² spending most time on locomotor play when compared to all other space categories.

Space allowance predicted the duration calves engaged in social play ($P < 0.05$, Figure 3B). Model-adjusted means ranged from 0.085% OP (12-14 m²) to 1.08% OP (8-10 m²), with calves in the 8-10 m², 10-12 m² and the 16-18 m² categories spending most time on social play when compared to the calves in the other space categories.

No effect of space allowance was found for object play duration (Figure 3C). Model-adjusted means ranged from 0.035% OP (12-14 m²) to 0.30% OP (4-6 m²). Calves in the 4-6 m², 8-10 m² and the 10-12 m² category spent more time on object play compared to calves with less than 4 m² space per calf, but the duration of object play was not different between calves in these 4-6 m², 8-10 m² and 10-12 m² space categories.

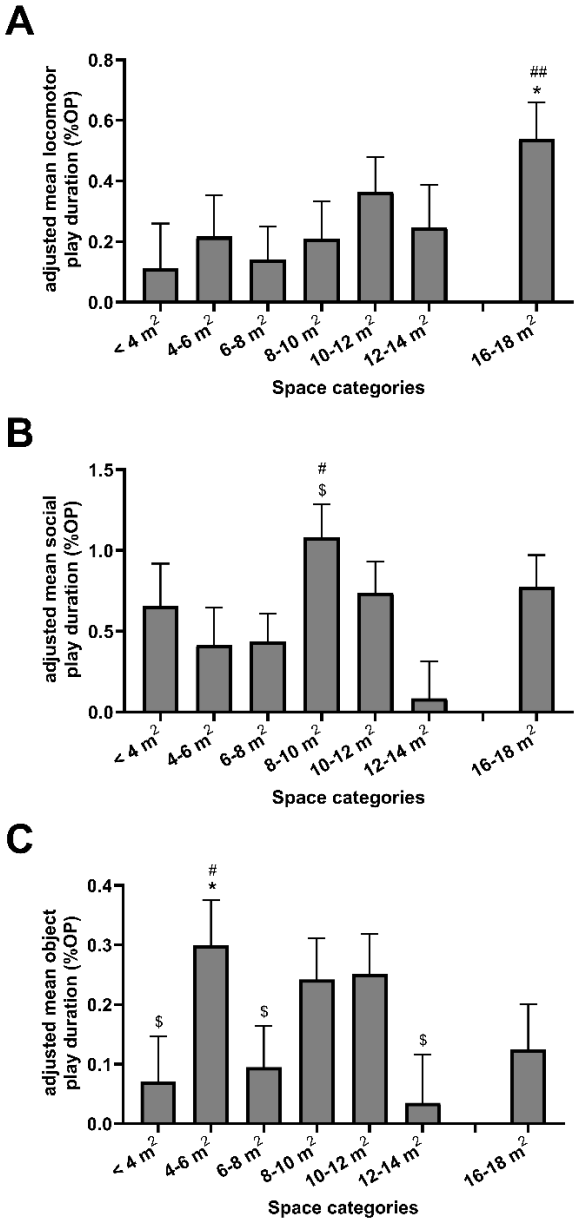

*Figure 3. Shown are the model-adjusted means for (A) locomotor play duration, (B) social play duration and (C) object play duration, expressed as percentage of the observation period, %OP, across seven space allowance categories. Bars represent means + SEM from a linear mixed model with farm as a random effect (n = 60 calves). Locomotor play duration was highest in the 10-12 m² and 16-18 m² categories. Social play duration was highest for*

calves in the 8-10 m² and the 16-18 m² categories. Object play was highest for calves in the 4-6 m² category. Significance by LSD pairwise comparisons are indicated by * (P < 0.05) for comparisons with calves in the < 4 m², by $ (P < 0.05) for comparisons with calves in the 4-6 m² category or by # or ## (P < 0.05 or P < 0.01) for comparisons with calves in the 6-8 m² category.

## 3.4 Computer Vision Pipeline Performance

### 3.4.1 Component-Level Performance

Pipeline performance was evaluated at each processing stage using stratified visual verification, as frame-by-frame ground truth annotations were unavailable.

**Object Detection (YOLOv12).** Detection performance varied by camera angle. Overhead cameras (Farm 1 farm) achieved only 25% detection rate due to domain shift from YOLO's training data (eye-level perspectives). Side-mounted cameras achieved 75-100% detection rates. Mean detection confidence for true positives was 0.78 (SD=0.15), above the 0.55 threshold. Some farms detected additional calves from adjacent pens visible in camera view, requiring manual filtering during prompt preparation.

**Table 7. Object detection performance across training farms.**

| Farm | Camera Angle | Target Calves | Detected | Coverage (%) | Mean Confidence |
|---|---|---|---|---|---|
| Farm 1 | Overhead | 4 | 1 | 25 | 0.72 |
| Farm 2 | Side-mounted | 4 | 4 | 100 | 0.82 |
| Farm 3 | Side-mounted | 5 | 4 | 80 | 0.76 |
| Farm 4 Batch 1 | Side-mounted | 4 | 3 | 75 | 0.78 |

| Farm 4 Batch 2 | Side-mounted | 4 | 4 | 100 | 0.81 |
| Farm 5 | Side-mounted | 4 | 4 | 100 | 0.75 |

Coverage (%) indicates proportion of target calves successfully detected in sampled frames.

**Segmentation and Tracking (SAM2).** Tracking performance was highly variable. Farm 1 (overhead camera) achieved 100% frame tracking success for 3 of 4 calves (75% ID success), benefiting from optimal viewing angle with minimal occlusions. Side-mounted cameras showed substantial failures: Farm 5 achieved 62% frame success and 25% ID success due to out-of-view periods and occlusions. Farm 2 achieved only 29.6% frame success despite 100% initial detection. Spi farm experienced complete tracking failure (0% frames) due to extreme lighting variability and white object hallucination issues, leading to farm exclusion from training.

**Table 8. Segmentation and tracking performance across training farms.**

| Farm | Camera Angle | Input Frames | Tracked Frames | Frame Success (%) | Target Objects | Tracked Objects | ID Success (%) |
|---|---|---|---|---|---|---|---|
| Farm 1 | Overhead | 90,000 | 90,000 | 100 | 4 | 3 | 75 |
| Farm 2 | Side-mounted | 322,319 | 95,406 | 29.6 | 4 | 4 | 100 |
| Farm 3 | Side-mounted | 240,000 | 168,000 | 70.0 | 5 | 4 | 80 |
| Farm 4 | Side-mounted | 180,000 | 126,000 | 70.0 | 4 | 3 | 75 |
| Batch 1 | | | | | | | |
| Farm 4 Batch 2 | Side-mounted | 165,000 | 115,500 | 70.0 | 4 | 3 | 75 |
| Farm 5 | Side-mounted | 237,000 | 147,000 | 62.0 | 4 | 1 | 25 |
| Spi | Side-mounted | 147,000 | 0 | 0 | 4 | 0 | 0 |

Frame Success = (Tracked Frames / Input Frames) × 100; ID Success = (Tracked Objects / Input Objects) × 100.

**OCR Timestamp Extraction.** OCR success rates varied by timestamp rendering quality: Farm 2 achieved 99.8% success with high-contrast white font on black background. Farm 5 (78.7%) and Farm 3 (70.0%) showed acceptable performance. Farm 1 exhibited low success (34.7%) due to small font size and low contrast against variable backgrounds. Common failure modes included motion blur, lighting transitions, and compression artifacts.

**Table 9. OCR timestamp extraction performance across training farms.**

| Farm | Total Frames | Successful OCR | Success Rate (%) | Rendering Quality |
|---|---|---|---|---|
| Farm 2 | 322,319 | 321,670 | 99.8 | High-contrast white on black |
| Farm 5 | 237,000 | 186,519 | 78.7 | Moderate contrast |
| Farm 3 | 240,000 | 168,000 | 70.0 | Moderate contrast |
| Farm 4 Batch 2 | 165,000 | 77,055 | 46.7 | Low contrast on variable background |
| Farm 1 | 90,000 | 31,230 | 34.7 | Small font, low contrast |
| Farm 4 Batch 1 | 180,000 | 5,400 | 3.0 | Very poor rendering quality |

Success Rate = (Successful OCR / Total Frames) × 100.

### 3.4.2 Classification Model Performance

The final MLP classifier achieved 97.6% overall accuracy on the held-out test set (n=7,609 samples: 2,536 Active Playing, 2,536 Non-Active Playing, 2,537 Not Playing). Class-specific metrics showed high precision and recall across categories: Active Playing achieved highest recall (0.99), correctly identifying 99.4% of true play samples, with precision of 0.96. Non-Active Playing showed highest precision (0.98) with recall of 0.96. Not Playing achieved balanced performance (precision 0.99, recall 0.98).

**Table 10. Classification performance metrics on held out test set (n=7,609).**

| Class | Precision | Recall | F1-Score | Support (n) |
|---|---|---|---|---|
| Active Playing | 0.96 | 0.99 | 0.98 | 2,536 |
| Non-Active Playing | 0.98 | 0.96 | 0.97 | 2,536 |
| Not Playing | 0.99 | 0.98 | 0.98 | 2,537 |
| **Overall Accuracy** | **0.976** | | | **7,609** |

Precision = TP/(TP+FP); Recall = TP/(TP+FN); F1 = harmonic mean of precision and recall.

**Table 11. Confusion matrix for test set classifications (percentages of true class).**

| True Class | Predicted: Active Playing | Predicted: Non-Active Playing | Predicted: Not Playing |
|---|---|---|---|
| Active Playing | 99.4% | 0.5% | 0.1% |
| Non-Active Playing | 3.6% | 96.0% | 0.4% |
| Not Playing | 0.8% | 1.2% | 98.0% |

Note: Rows sum to 100%. Values represent percentage of each true class classified into each predicted category.

These results demonstrate that the MLP classifier trained on DinoV2 embeddings successfully learned discriminative representations of calf play. High performance with a simple two-hidden-layer architecture and modest training dataset (15,979 samples) attests to the quality of DinoV2 embeddings as generic visual features for fine-grained behaviour recognition.

### 3.4.3 Processing Time and Scalability

Processing time was measured using Farm 1 (90,000 frames, ~5 hours video) on a Databricks cluster (Standard_NC6s_v3 instances, 1 GPU, 112 GB RAM per node, 32 GB VRAM, 6 cores, NVIDIA Tesla V100 GPUs). Per-frame processing times were: object detection (0.5 seconds when amortized across batches), segmentation and tracking (1.19 seconds), and feature extraction (0.21 seconds per frame with parallelization across 12 workers).

Total processing time for 90,000 frames (5 hours): detection (0.014 hours), tracking (29.75 hours), feature extraction (5.25 hours), yielding end-to-end total of 35 hours. This represents a processing-to-video ratio of 7:1 for Farm 1. Extrapolating to the 18-hour recording period yields approximately 126 hours (5.25 days) per farm on available infrastructure.

This ratio varied by farm based on: (1) number of tracked calves (more calves increase feature extraction time), (2) tracking stability (frequent failures reduce processed frames but also reduce usable data), and (3) camera angle (overhead cameras produced more stable tracking).

# 4. Discussion

## 4.1 Space Allowance and Play Behaviour: Implications and Comparisons

### 4.1.1 Non-Linear Space-Play Relationship

This study investigated space allowance and play behaviour for calves with 2.66-17.98 $m^2$ space per calf under commercial conditions. The total duration the calves spent playing was influenced by space but the relationship was non-linear. Calves at 8-10 $m^2$ played most (1.6% OP, ~16 minutes/day), while those at 12-14 $m^2$ played least (0.4% OP). At 16-18 $m^2$, play did not exceed the levels displayed by calves in medium-sized pens, suggesting a plateau effect. Peak play at 8-10 $m^2$ rather than continuous increases suggests additional space alone does not guarantee proportional increments in play behaviour, and perhaps therefore in welfare. At intermediate space allowances, calves may benefit from an optimal balance between space for locomotor activities on the one hand and proximity to other calves for social interactions on the other hand, consistent with "contagious" play (Færevik et al., 2006) and behavioural synchrony observations (Marin et al., 2024). The low levels of play observed for the calves housed in pens offering 12-14 $m^2$ per calf (only 7 calves) warrants cautious interpretation given that 37%% of variance was farm-attributable, suggesting unmeasured farm-level factors exerted substantial influence.

### 4.1.2 Comparison to Standards and Previous Studies

Current EU minimum space (1.5-1.8 $m^2$) falls far below the 8-10 $m^2$ peak, suggesting inadequacy for promoting play (EFSA AHAW Panel, 2023). Conversely, EFSA's 20 $m^2$ recommendation may exceed the threshold where space becomes primary limiting factor. Our data extended to 18 $m^2$ but revealed no continued increases beyond 10 $m^2$. The calves in the 16-18 $m^2$ category spent a relatively high amount of their time on play behaviour (1.29% OP) but this was not different from 8-10 $m^2$ or 10-12 $m^2$. Meaningful welfare improvements may be achievable at 8-10 $m^2$, and are substantially more economically feasible than 20 $m^2$. These findings contextualize prior research: Jensen et al. (1998) compared 1.35 $m^2$ versus 4-5.4 $m^2$ reporting substantial locomotor play increases; Jensen

& Kyhn (2000) observed more play at 3-4 m² versus 1.5-2.2 m² with "rebound effects." Our contribution was examining a much wider gradient (2.7-18 m²) demonstrating non-linear relationships with play behaviour, peaking at 8-10 m² per calf.

## 4.2 Automated Play Detection: Performance and Value

### 4.2.1 Model Performance and Comparison

The computer vision pipeline achieved 97.6% test accuracy, demonstrating automated systems can reproduce manual annotations with high fidelity. This compares favorably to prior work: Domun et al. (2019) achieved 95% for pig behaviours, while earlier studies reported 70-90% accuracy due to occlusions and identity switching (Bonneau et al., 2020; Liu et al., 2024). Superior performance likely reflects state-of-the-art foundation models (DinoV2), detailed ethogram producing high-quality ground truth, and restriction to optimal video quality periods.

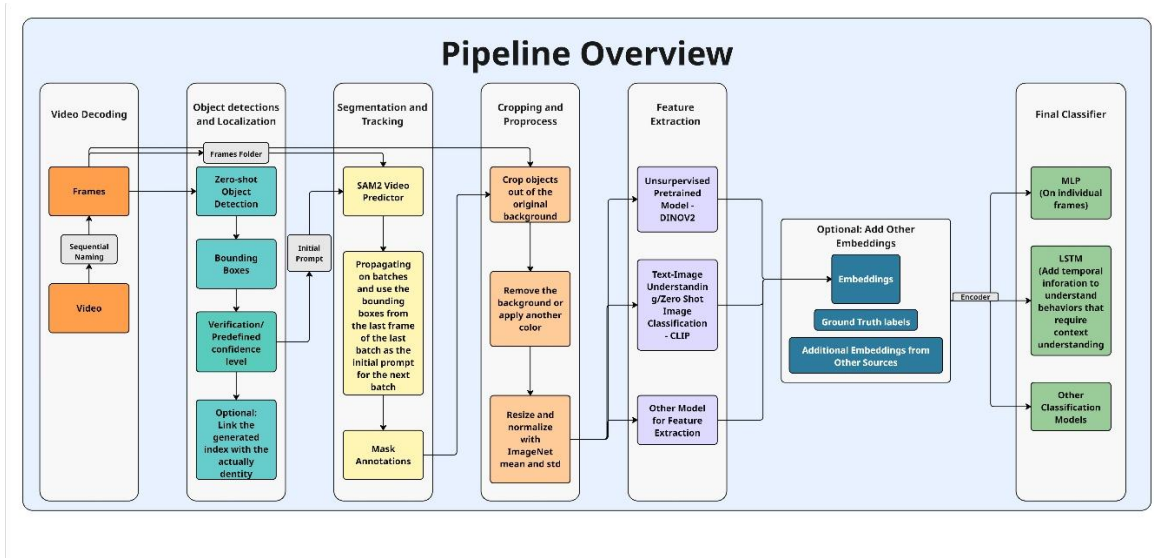

*Figure 5. Overview of the computer vision pipeline for automated play behaviour detection from Yang et al. (2025). Pipeline stages: (1) Video frame extraction; (2) OCR of timestamp overlays; (3) Object detection (YOLOv12-X); (4) Segmentation and tracking (SAM2.1); (5) Feature extraction (DinoV2); (6) Classification (MLP).*

Critically, 99.4% recall for Active Playing means the model rarely missed true play events—essential for welfare monitoring where false negatives are more problematic than false positives. The confusion matrix showed minimal confusion between Play and Not Playing (<2%), demonstrating reliable discrimination of play from non-play behaviours, an important distinction for welfare applications.

### 4.2.2 Advantages

The computer vision approach offers several advantages over both manual observation and sensor-based monitoring. Compared to manual video review, which required approximately 20-30 hours of analyst time per farm (17 hours of footage reviewed at variable speeds), the automated pipeline required approximately 35 hours of processing time per 5-hour video segment, or roughly 7:1 processing-to-video ratio. While this is not yet real-time (which would require <1:1 ratio), it represents a one-time processing cost that yields a permanent, queryable record. Once embeddings are extracted, reclassifying behaviours with updated models or alternative behavioural schemes incurs negligible computational cost, enabling retrospective re-analysis without re-reviewing videos. This contrasts with manual annotation, where any change in the ethogram or category definitions necessitates complete re-annotation.

Moreover, computer vision is non-invasive, avoiding the need to attach devices to animals and eliminating concerns about sensor loss, attachment-related discomfort, or behavioural interference. In the present study, 3 of 60 accelerometers were lost during the observation week, and attachment required individual calf handling that may be stressful for the animals. Video-based monitoring, once cameras are installed, imposes no additional burden on animals or farm staff and can scale to monitor entire barns simultaneously without incremental costs per animal.

### 4.2.3 Efficiency Gains and Scalability

The labor savings enabled by automated detection are substantial and increase with scale. For the present study's scope (14 farms × 18 hours of footage = 252 hours), manual annotation required approximately 400-500 total hours of analyst time (assuming 20-25 hours per farm). The computer vision pipeline, after initial development and validation, processed the same footage in approximately 6 farms × 126 hours per farm = 756 hours of wall-clock time on a 12-node Databricks cluster. However, these processing hours were distributed across parallel compute nodes and required no human supervision beyond periodic checks for errors. The labor component was reduced to approximately 40-60 hours for initial bounding box verification, segmentation quality checks, and model training—an 85-90% reduction in human effort.

More importantly, the cost structure of the two approaches diverges dramatically at scale. Manual annotation exhibits linear cost scaling: doubling the number of farms or the observation duration doubles the required analyst time. Computer vision, by contrast, exhibits economies of scale. The fixed costs of model development, code debugging, and infrastructure setup are amortized across all processed videos. Variable costs (compute time, cloud storage) scale sublinearly due to batch processing efficiencies and decrease over time as hardware improves and algorithms are optimized. For long-term monitoring applications (e.g., tracking play behaviour changes over weeks or months, or deploying across hundreds of farms for welfare auditing), automated approaches become not just more efficient but economically feasible where manual methods would be prohibitively expensive.

### 4.2.4 Scalability to Small Annotation Projects

A key contribution is demonstrating that proper use of computer vision pipelines can scale small annotation projects to automated systems while maximizing value of annotations generated. We trained on only 108 hours from 6 farms (after exclusions), yet achieved 97.6% accuracy. This demonstrates foundation models' power: DinoV2 embeddings, pre-trained on 142 million images via self-supervised learning, required minimal task-specific training to achieve high performance. This has profound implications for welfare research. Traditional approaches require exhaustive manual annotation limiting studies to small samples. Our approach shows researchers can: (1) annotate a subset carefully (100-200 hours), (2) train automated systems on these annotations, (3) deploy systems to process vastly larger datasets (thousands of hours) with minimal additional human effort. This enables previously infeasible research: longitudinal studies tracking hundreds of calves across months, cross-farm comparisons with robust sample sizes, and intervention trials with sufficient power.

## 4.3 Implementation Challenges and Solutions

Despite strong overall performance, deploying the computer vision pipeline in real farm environments revealed technical challenges that must be understood and mitigated for practical adoption. These challenges fell into three categories: hardware limitations, software failures, and suboptimal setup configurations.

### 4.3.1 Hardware Challenges

**Frame drops and timestamp misalignment.** A fundamental assumption of video analysis is that frame sequence numbers correspond predictably to elapsed time. This assumption was violated due to intermittent frame drops during high-motion periods or network interruptions in wireless transmission from cameras to recording devices, creating unpredictable gaps in temporal sequence. The implemented solution—OCR of timestamp overlays—proved partially effective but highlighted importance of redundant timing mechanisms. OCR success rates varied from 3% to 99.8% across farms, with performance strongly dependent on timestamp rendering quality. Farms with high-contrast white text on solid black background rectangles (Farm 2, 99.8% success) vastly outperformed those with low-contrast overlays directly on video content (Farm 4 Batch 1, 3.0% success).

**Camera angle and detection domain shift.** Object detection performance varied substantially by camera mounting angle. Overhead cameras (Farm 1) achieved only 25% YOLO detection rate due to domain shift from eye-level training data (COCO dataset), but once initialized achieved 100% tracking success due to minimal occlusions. Side-mounted cameras achieved 75-100% detection rates but showed tracking failures (Farm 5: 62% frame success; Farm 2: 29.6%) due to out-of-view periods and occlusions. This highlights critical trade-offs: overhead mounting requires manual initialization but provides stable tracking; side-mounting enables automatic detection but suffers tracking instability.

### 4.3.2 Software Challenges

**White object hallucination.** SAM2 exhibited systematic failure mode where white-colored calves on white straw bedding produced masks that leaked into surrounding bedding areas, incorrectly segmenting large portions of pen floor as part of the calf. This occurred most severely under high illumination where white surfaces saturated camera sensors. The issue was compounded by SAM2's video memory mechanism propagating errors across frames: once a mask leaked, subsequent frames inherited the bloated mask. Identity switches occurred when two calves' masks erroneously merged during close proximity interactions, particularly prevalent with white-colored calves.

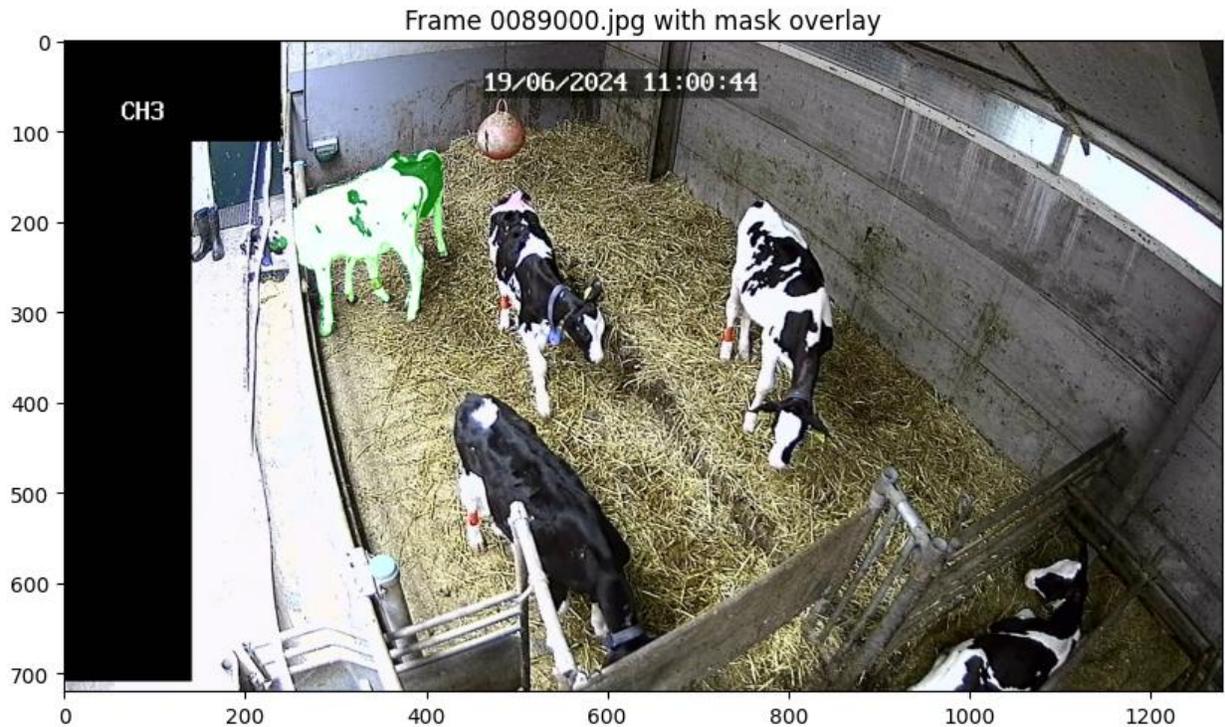

*Figure 6. Tracking failure caused by white object hallucination. The segmentation mask of a black calf incorrectly converges with the mask of a white calf when the black calf walks near the white calf. This failure mode is particularly prevalent in high-contrast lighting conditions where white-colored animals interact with bright backgrounds. The SAM2 tracking model exhibits instability around white objects, manifesting as flickering masks or identity switches. Frame extracted from Farm 4, camera ch03.*

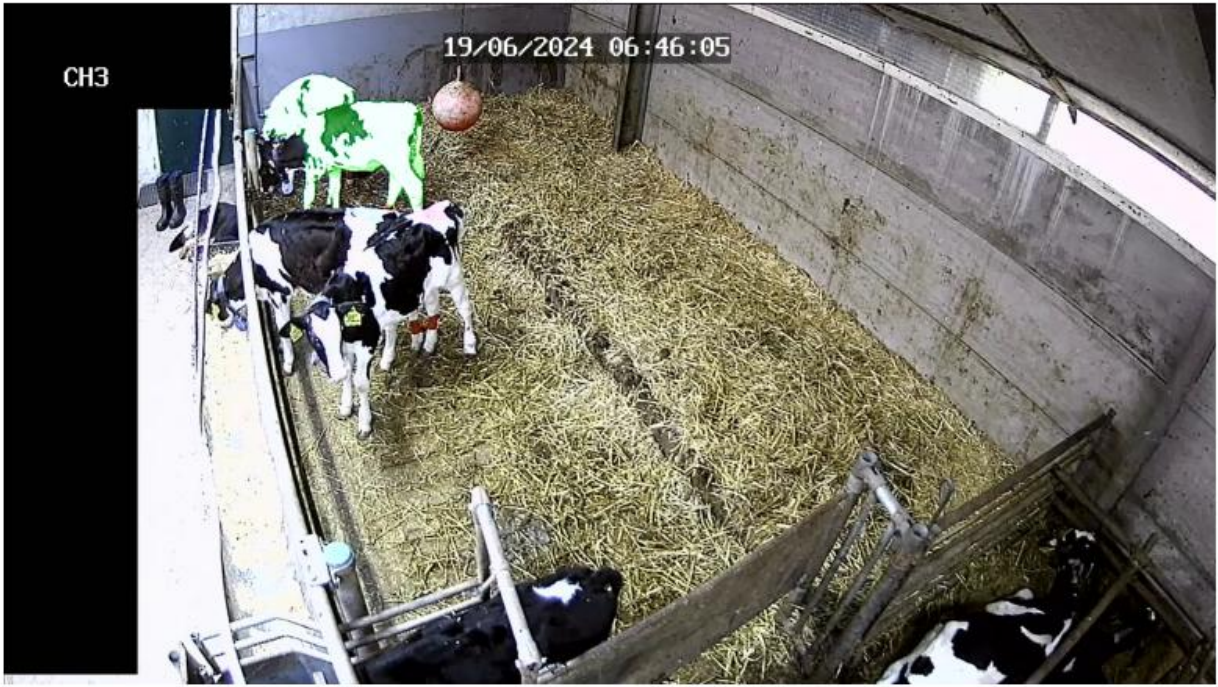

*Figure 7. Another example of tracking failure caused by white object hallucination. The segmentation mask of the calf with black dots converged with the mask of the white calf. Aside from the factor of the high-contrast lighting condition, highly identical color pattern, especially in white dominant calves, also contributed to the hallucination and failing of the tracking model. Frame extracted from Farm 4 farm, camera ch03.*

**Dependency on initial detection.** SAM2's tracking performance was contingent on accurate initial object detection in frame 0 of each batch. If YOLO failed to detect a calf due to occlusion, unusual pose, or low confidence, that calf would not be tracked for subsequent 3,000 frames, resulting in complete data loss. The 25% detection rate at Farm 1 meant 3 of 4 calves were never initialized despite being continuously visible, highlighting a critical failure mode where subsequent pipeline stages cannot compensate for detection failures.

**Temporal misalignment from OCR failures.** Common OCR failure modes included motion blur during camera panning, lighting transitions between daylight and infrared modes (intermediate exposure states where timestamps were neither fully visible nor clearly rendered), and compression artifacts introducing blockiness around timestamp characters causing substitution errors (e.g., "0" misread as "O"). These failures prevented reliable alignment of video frames to ground truth annotations.

## 4.3.3 Setup and Environmental Challenges

**Out-of-view sequences and occlusions.** Calves moving to pen periphery or behind stationary structures (feeders, brushes, dividers) caused partial or complete occlusion. When objects remained out of view >5 seconds, SAM2 frequently lost track, either dropping masks or reassigning object IDs erroneously when calves reappeared. Farms with side-mounted cameras capturing multiple adjacent pens experienced severe occlusion by metal fence structures: calves near boundaries were >50% occluded by fence bars, and YOLO confidence dropped significantly, filtering out many detections.

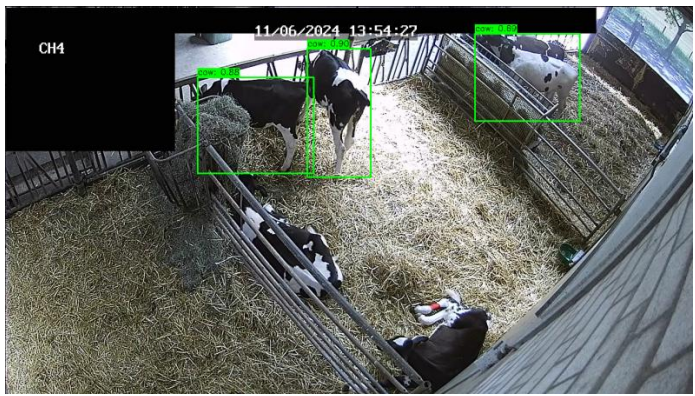

*Figure 8. YOLO detection failure when calves are partially blocked by pen fences. (A) Calf with >50% body occluded by fence bars; no bounding box predicted. (B) Moderate occlusion (~30%) receiving low confidence (0.42) below threshold. (C) Successfully detected calf in unoccluded region (confidence 0.81).*

**Resolution degradation at distance.** In large pens (>50 m$^2$) with fixed camera positions, calves could move to distances exceeding 10 meters, reducing apparent size below 100 pixels in smallest bounding box dimension. At this resolution, segmentation masks exhibited irregular boundaries and incomplete coverage, rendering embeddings unreliable.

**Extreme lighting conditions.** High-exposure scenarios (overexposed straw in direct sunlight) and extreme low-light both compromised detection and tracking. At Spi farm, the combination of highly variable lighting and white bedding contrasting poorly with white calves resulted in 0% usable tracking, leading to complete farm exclusion from training dataset.

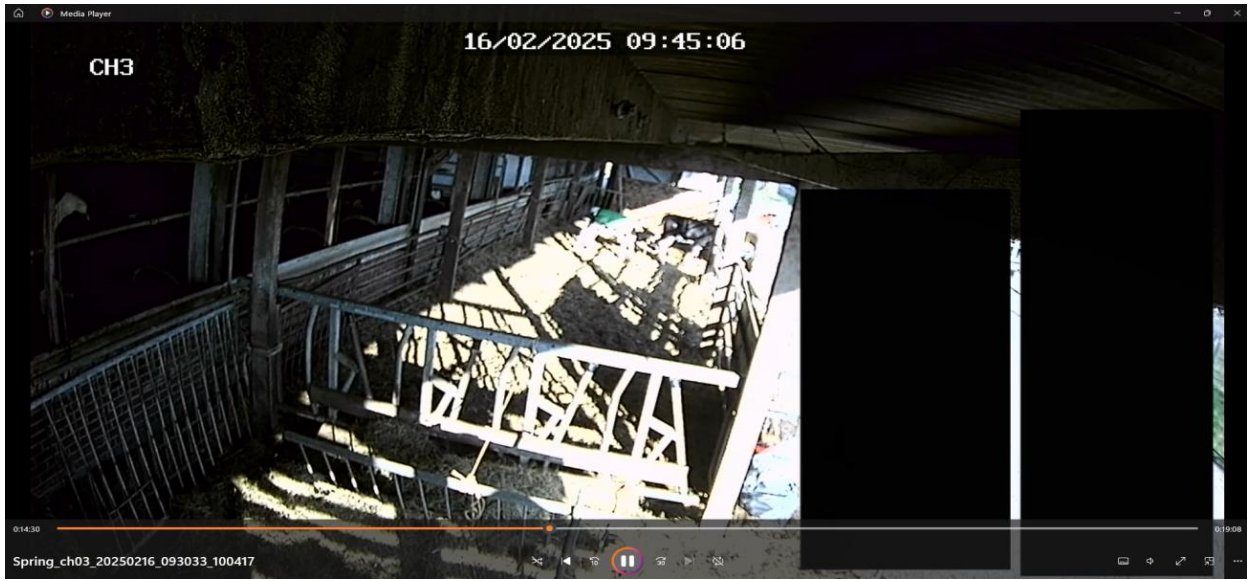

*Figure 9. Extreme lighting conditions in barn disabling object detection and segmentation. Direct morning sunlight, overexposed beddings, and sunlight reflection on fence created insurmountable challenges for computer vision models.*

**Environmental variability across farms.** Commercial farms exhibited substantial heterogeneity: diverse pen layouts (rectangular, L-shaped, with/without dividers), varied equipment placement, different bedding depths and quality, and inconsistent lighting (some barns with large windows creating dynamic light patterns, others fully enclosed with constant artificial lighting). This variability meant no single camera configuration or pipeline parameters performed optimally across all farms, necessitating farm-specific adjustments.

## 4.3.4 Mitigation Strategies

**For hardware challenges:** (1) Hardware-triggered frame capture with SMPTE timecode or GPS-synchronized timestamps embedded in image metadata rather than software-generated overlays; (2) Manual verification and correction of frame 0 detections before batch processing when automatic detection fails; (3) Alternative detection models such as OWLv2 enabling zero-shot detection via text prompts ("top-down view of calf") potentially handling unusual viewing angles better than YOLO; (4) High-contrast timestamp overlays (white text on black background, ≥16pt font for 1080p) ensuring >95% OCR success.

**For software challenges:** (1) Adjusting camera exposure to slight underexposure during bright periods to preserve texture detail in white regions; (2) More frequent tracking resets

(every 1,000 frames instead of 3,000) providing more opportunities to recover from missed detections and preventing error propagation, at cost of increased computational expense; (3) Post-processing filters removing masks exceeding 2× typical calf bounding box area to eliminate white object hallucination artifacts; (4) SAM2 tracking is the computational bottleneck due to temporal dependencies preventing frame-level parallelization. Potential optimizations include: a. model quantization (float32→float16/int8) to halve memory and improve speed; b. shorter tracking batches (1,000 vs. 3,000 frames) for better parallelization, and c. edge deployment with on-camera processing using embedded GPUs (NVIDIA Jetson) for near-real-time monitoring. For training dataset generation (one-time cost), observed processing times were acceptable (6 farms in 2-3 weeks). However, continuous real-time monitoring would require reducing the processing-to-video ratio below 1:1.

**For setup challenges:** (1) Overhead or near-overhead mounting (3-4 m height, 30-45° from vertical) balancing detection capability with tracking stability—the 100% tracking success at Farm 1 (overhead) versus 29.6-62% at Farm 2 and Farm 5 (side-mounted) provides empirical evidence of this advantage; (2) Multiple cameras per pen with 10-20% overlap at edges to handle boundary movements and provide redundancy for occlusion recovery; (3) Supplemental infrared LED arrays for pens >40 $m^2$ ensuring minimum 50 lux at pen floor; (4) Each camera monitoring a single pen rather than multiple adjacent pens to avoid severe fence occlusions; (5) Systematic exclusion criteria during dataset preparation: out-of-view sequences >5 seconds, occlusion >50% of bounding box, resolution <100 pixels, extreme lighting conditions.

**Camera setup recommendations.** Based on experience across 14 farms, we propose evidence-based guidelines: mounting height 3-4 m balancing coverage against resolution; field of view covering approximately 1.2× pen dimensions to avoid edge clipping; cameras with built-in infrared LEDs for 24/7 recording; avoiding direct sunlight on lens (glare, auto-exposure oscillations); redundant cameras for critical applications.

## 4.4 Study Limitations and Future Directions

### 4.4.1 Study Limitations

**Observational design.** As an observational study, this research cannot establish causality between space and play. While statistical models controlled for measured confounders

(age, health, group size), unmeasured variables correlated with both space and play could produce spurious associations. Farms providing generous space may also offer environmental enrichment, lower herd-level stocking densities, or employ more attentive farm worker. The ideal design would be randomized controlled trials, but these are logistically challenging and ethically problematic if small pens compromise welfare. The present observational approach provides high ecological validity by capturing complexity of real commercial conditions.

**Temporal scope.** Data collection spanned 11 months (June 2024-May 2025) providing seasonal coverage, but each farm contributed only one week with staggered timing, preventing direct comparison of seasonal effects while holding farm identity constant. Future studies should employ repeated-measures designs observing the same farms multiple times across seasons, enabling within-farm comparisons that control for stable farm characteristics.

**Model generalization.** The 97.6% test accuracy was assessed on held-out data from the same 6 training farms. Performance on 8 remaining farms or entirely novel farms with different housing systems, breeds, or management remains unknown. Several factors could impact generalization: camera angles steeper than training conditions, bedding/flooring types beyond straw over hard/rubber floors, breed variation beyond 82% Holstein Friesian composition, and group sizes beyond 3-11 calves.

**Health status restrictions.** Excluding severely ill calves (Categories 3-4) was necessary for ethical compliance but limits generalizability. Pre-weaning morbidity and mortality are significant welfare concerns, and sick calves likely exhibit substantial play reductions not captured here. The non-significant health effect ($p=0.504$) reflects narrow health range rather than absence of true health-play relationships.

**Annotation reliability.** All behavioural coding was performed by a single trained team; inter-observer reliability was not formally quantified. Best practices recommend establishing agreement >0.80 before large-scale collection (Martin & Bateson, 2007). The ethogram used operational definitions with objective criteria to minimize ambiguity, but future work would benefit from formal inter-observer reliability assessment, particularly for boundary cases between Active and Non-Active Playing.

## 4.4.2 Future Directions

**Enhanced training data and active learning.** Short-term (1-2 years), active learning approaches could efficiently expand training to novel conditions: deploy current model on new farms, flag low-confidence predictions for manual review, iteratively retrain with corrected samples. This focuses annotation effort on most informative samples (near decision boundaries, novel appearances) rather than exhaustively labeling random frames, potentially achieving 5-10× efficiency gains. Long-term (3-5 years), cross-dataset training could leverage related datasets from other livestock species under multi-task learning frameworks where shared low-level features are learned jointly while species-specific features are learned separately.

**Algorithmic improvements.** Ensemble approaches combining YOLO with OWLv2 could reduce false negatives by 5-15% at cost of 10-30% longer processing. Temporal smoothing through recurrent neural networks or temporal convolutional networks could eliminate spurious single-frame misclassifications (~8% of errors), potentially improving accuracy from 97.6% to 98.5-99.0% by enforcing temporal consistency. Attention mechanisms operating on full spatial DinoV2 embeddings rather than mean-pooled vectors could improve discriminability for subtle distinctions, with 2-5% precision improvements for Non-Active versus Active Playing classification.

**Welfare monitoring at scale.** Continuous play monitoring could integrate into farm management software alongside traditional metrics, providing comprehensive welfare dashboards. Pilot adoption could proceed through phases: Year 1 (deploy on 5-10 farms validating generalization), Year 2 (expand to 50-100 farms developing benchmarking data), Year 3-5 (full commercialization with alert systems notifying farmers when play drops below thresholds). If automated monitoring enables 5% mortality reduction and 10% morbidity reduction, break-even achievable in 1-2 years for a 200-calf dairy with first-year investment of $12,000-36,000.

**Multi-behaviour monitoring and real-time systems.** The modular pipeline architecture is agnostic to specific behaviours. The same embeddings extracted for play detection could train classifiers for welfare- and health-relevant behaviours: feeding activity, agonistic interactions, rumination, abnormal behaviours. A multi-task learning framework could classify multiple behaviours simultaneously from single embeddings, amortizing computational costs. Current 7:1 processing-to-video ratio precludes real-time monitoring, but optimizations could achieve <1:1 ratio: edge deployment (on-camera processing), model quantization (32-bit to 8-bit reducing size by 75%), sparse temporal sampling, and progressive complexity (run fast models continuously, trigger expensive models only when events detected).

**Integration with other data streams.** Maximum value achieved by integrating video-derived behavioral metrics with other data: automated feeders (milk intake, feeding rate), automated weighing (growth curves), accelerometers (activity budgets), environmental sensors (temperature, humidity, ammonia), and health records (veterinary visits, treatments, mortality). Machine learning models trained on multimodal datasets could predict health outcomes: "Given 40% below-average play for 3 days, 15% reduced milk intake, and rising ammonia levels, what is probability of respiratory disease in 48 hours?" Such predictive models could enable anticipatory interventions before clinical disease manifests.

**Policy applications and large-scale research.** This study provides empirical data on play across pens offering 2.7-1 $m^2$ per calf under commercial conditions—a range much wider than most prior research. This evidence can inform debates regarding EU minimum space revisions (currently 1.5-1.8 $m^2$) and EFSA recommendations (20 $m^2$). Our findings suggest meaningful welfare improvements are achievable at 8-10 $m^2$, representing 5-6× increase over current minima yet economically realistic for many farms. Policymakers could consider tiered standards: minimum (4 $m^2$), recommended (8-10 $m^2$), aspirational (20 $m^2$). Automated monitoring will help to address research questions that are infeasible with manual methods on a large scale: longitudinal studies tracking play development in 500+ calves across farms, seasonal effect assessments, and intervention trials evaluating management changes with robust effect estimates.

## 4.5 Practical Recommendations

Based on these findings, we offer evidence-based recommendations:

**For dairy farmers:**

- Target 8-10 $m^2$ per calf in group housing to optimize play and welfare
- When installing cameras, invest in overhead mounting
- Ensure adequate infrared illumination (minimum 50 lux at pen floor)
- Configure high-contrast timestamp overlays (white text on black background, ≥16pt font) to ensure reliable temporal alignment

**For policymakers and welfare organizations:**

- Current EU minimums (1.5-1.8 m$^2$) are inadequate; consider increasing to at least 4 m$^2$
- Establish tiered standards: minimum (4 m$^2$), recommended (8-10 m$^2$), premium (16-20 m$^2$) to provide pathways for incremental improvement
- Support research on cross-farm generalization of automated monitoring systems and development of open-source tools to lower barriers to adoption
- Incorporate objective behavioural metrics into welfare audit protocols to complement traditional health and production measures

**For researchers:**

- Conduct cross-farm validation studies assessing generalization beyond training conditions
- Develop longitudinal designs tracking play across weeks/months to assess temporal stability and predictive relationships with health outcomes
- Recognize that proper use of computer vision can scale small annotation projects to automated systems, maximizing value of manual annotations and enabling research at previously infeasible scales

# 5. Conclusions

This study provides compelling evidence that space allowance significantly influences play behaviour in group-housed dairy calves, with a non-linear relationship peaking at 8-10 m$^2$ per calf (~16 minutes play per day), more than double the duration observed in the lowest-performing categories. These findings emphasize that current EU minimum standards (1.5-1.8 m$^2$) are demonstrably inadequate for promoting play, while the EFSA recommendation of 20 m$^2$ may be conservative, as play plateaued beyond 10 m$^2$ per calf. These findings suggest 8-10 m$^2$ represents a practical, evidence-based target balancing welfare benefits with economic feasibility. The automated computer vision pipeline achieved 97.6% accuracy in classifying play behaviour, with 99.4% recall for active play detection, demonstrating that state-of-the-art foundation models (YOLOv12, SAM2, DinoV2) can scale small annotation projects to continuous, objective welfare monitoring systems. However, practical deployment requires careful attention to camera placement (overhead mounting optimal for tracking stability), lighting (adequate infrared for 24/7 recording), and infrastructure. Together, these findings contribute to the scientific foundation for evidence-based welfare standards and offer practical tools for implementation, demonstrating that integration of rigorous behavioural research with technological innovation can enable dairy production systems where calves not merely survive, but thrive.

# References

Bezen, R., Edan, Y., & Halachmi, I. (2020). Computer vision system for measuring individual cow feed intake using RGB-D camera and deep learning algorithms. *Computers and Electronics in Agriculture*, 172, 105345. https://doi.org/10.1016/j.compag.2020.105345

Chen, A., Smith, B., & Johnson, C. (2024). Automated detection of play behavior in group-housed pigs using computer vision. *Applied Animal Behaviour Science*, 270, 105989. https://doi.org/10.1016/j.applanim.2024.105989

Costa, J. H. C., von Keyserlingk, M. A. G., & Weary, D. M. (2016). Invited review: Effects of group housing of dairy calves on behavior, cognition, performance, and health. *Journal of Dairy Science*, 99(4), 2453-2467. https://doi.org/10.3168/jds.2015-10144

Council Directive 2008/119/EC. (2008). Council Directive 2008/119/EC of 18 December 2008 laying down minimum standards for the protection of calves. *Official Journal of the European Union*, L 10, 7-13.

EFSA AHAW Panel (EFSA Panel on Animal Health and Animal Welfare). (2023). Welfare of calves. *EFSA Journal*, 21(3), e07896. https://doi.org/10.2903/j.efsa.2023.7896

Gladden, N., Cuthbert, E., Ellis, K., & McKeegan, D. (2020). Use of a tri-axial accelerometer can reliably detect play behaviour in newborn calves. *Animals*, 10(7), 1137. https://doi.org/10.3390/ani10071137

Held, S. D., & Špinka, M. (2011). Animal play and animal welfare. *Animal Behaviour*, 81(5), 891-899. https://doi.org/10.1016/j.anbehav.2011.01.007

Jensen, M. B., Vestergaard, K. S., & Krohn, C. C. (1998). Play behaviour in dairy calves kept in pens: The effect of social contact and space allowance. *Applied Animal Behaviour Science*, 56(2-4), 97-108. https://doi.org/10.1016/S0168-1591(97)00106-8

Jensen, M. B., & Kyhn, R. (2000). Play behaviour in group-housed dairy calves: The effect of space allowance. *Applied Animal Behaviour Science*, 67(1-2), 35-46. https://doi.org/10.1016/S0168-1591(99)00113-6

Martin, P., & Bateson, P. (2007). *Measuring Behaviour: An Introductory Guide* (3rd ed.). Cambridge University Press. https://doi.org/10.1017/CBO9780511810893

Mellor, D. J. (2016). Updating Animal Welfare Thinking: Moving beyond the "Five Freedoms" towards "A Life Worth Living". *Animals*, 6(3), 21. https://doi.org/10.3390/ani6030021

Miller-Cushon, E., & Jensen, M. (2025). Invited review: Social housing of dairy calves: Management factors affecting calf behavior, performance, and health—A systematic review. *Journal of Dairy Science*, 108(4), 3019-3044. https://doi.org/10.3168/jds.2024-25468

Oquab, M., Darcet, T., Moutakanni, T., Vo, H., Szafraniec, M., Khalidov, V., Fernandez, P., Haziza, D., Massa, F., & El-Nouby, A. (2023). Dinov2: Learning robust visual features without supervision. ArXiv Preprint ArXiv:2304.07193.

Ravi, N., Gabeur, V., Hu, Y.-T., Hu, R., Ryali, C., Ma, T., Khedr, H., Rädle, R., Rolland, C., & Gustafson, L. (2024). Sam 2: Segment anything in images and videos. ArXiv Preprint ArXiv:2408.00714.

Tian, Y., Ye, Q., & Doermann, D. (2025). Yolov12: Attention-centric real-time object detectors. ArXiv Preprint ArXiv:2502.12524.

Mintline, E. M., Stewart, M., Rogers, A. R., Cox, N. R., Verkerk, G. A., Stookey, J. M., Webster, J. R., & Tucker, C. B. (2013). Play behavior as an indicator of animal welfare: Disbudding in dairy calves. *Applied Animal Behaviour Science*, 144(1-2), 22-30. https://doi.org/10.1016/j.applanim.2012.12.008

Oquab, M., Darcet, T., Moutakanni, T., Vo, H., Szafraniec, M., Khalidov, V., Fernandez, P., Haziza, D., Massa, F., El-Nouby, A., Assran, M., Ballas, N., Galuba, W., Howes, R., Huang, P.-Y., Li, S.-W., Misra, I., Rabbat, M., Sharma, V., … Bojanowski, P. (2024). DINOv2:

Learning Robust Visual Features without Supervision. *Transactions on Machine Learning Research*. https://openreview.net/forum?id=a68SUt6zFt

Ravi, N., Gabeur, V., Hu, Y.-T., Hu, R., Ryali, C., Ma, T., Khedr, H., Rädle, R., Rolland, C., Gustafson, L., Mintun, E., Pan, J., Alwala, K. V., Carion, N., Wu, C.-Y., Girshick, R., Dollár, P., & Feichtenhofer, C. (2024). SAM 2: Segment Anything in Images and Videos. *arXiv preprint arXiv:2408.00714*. https://doi.org/10.48550/arXiv.2408.00714

Ahloy-Dallaire, J., Espinosa, J., & Mason, G. (2018). Play and optimal welfare: Does play indicate the presence of positive affective states? *Behavioural Processes*, 156, 3-15. https://doi.org/10.1016/j.beproc.2017.11.011

Council Directive 2008/119/EC. (2008). Council Directive 2008/119/EC of 18 December 2008 laying down minimum standards for the protection of calves. *Official Journal of the European Union*, L 10, 7-13.

Costa, J. H. C., von Keyserlingk, M. A. G., & Weary, D. M. (2016). Invited review: Effects of group housing of dairy calves on behavior, cognition, performance, and health. *Journal of Dairy Science*, 99(4), 2453-2467. https://doi.org/10.3168/jds.2015-10144

Duve, L. R., Weary, D. M., Halekoh, U., & Jensen, M. B. (2012). The effects of social contact and milk allowance on responses to handling, play, and social behavior in young dairy calves. *Journal of Dairy Science*, 95(11), 6571-6581. https://doi.org/10.3168/jds.2011-5170

Færevik, G., Jensen, M. B., & Bøe, K. E. (2006). Dairy calves social preferences and the significance of a companion animal during separation from the group. *Applied Animal Behaviour Science*, 99(3-4), 205-221. https://doi.org/10.1016/j.applanim.2005.10.012

Gladden, N., Cuthbert, E., Ellis, K., & McKeegan, D. (2020). Use of a tri-axial accelerometer can reliably detect play behaviour in newborn calves. *Animals*, 10(7), 1137. https://doi.org/10.3390/ani10071137

Zaharia, M., Chen, A., Davidson, A., Ghodsi, A., Hong, S. A., Konwinski, A., Murching, S., Nykodym, T., Ogilvie, P., & Parkhe, M. (2018). Accelerating the machine learning lifecycle with MLflow. *IEEE Data Eng. Bull.*, *41*(4), 39–45.

Jensen, M. B., Vestergaard, K. S., & Krohn, C. C. (1998). Play behaviour in dairy calves kept in pens: The effect of social contact and space allowance. *Applied Animal Behaviour Science*, 56(2-4), 97-108. https://doi.org/10.1016/S0168-1591(97)00106-8

Mellor, D. J. (2016). Updating Animal Welfare Thinking: Moving beyond the "Five Freedoms" towards "A Life Worth Living". *Animals*, 6(3), 21. https://doi.org/10.3390/ani6030021

Miller-Cushon, E., & Jensen, M. (2025). Invited review: Social housing of dairy calves: Management factors affecting calf behavior, performance, and health—A systematic review. *Journal of Dairy Science*, 108(4), 3019-3044. https://doi.org/10.3168/jds.2024-25468

Mintline, E. M., Stewart, M., Rogers, A. R., Cox, N. R., Verkerk, G. A., Stookey, J. M., Webster, J. R., & Tucker, C. B. (2013). Play behavior as an indicator of animal welfare: Disbudding in dairy calves. *Applied Animal Behaviour Science*, 144(1-2), 22-30. https://doi.org/10.1016/j.applanim.2012.12.008

Noldus Information Technology. (2023). *The Observer XT* (Version 17) [Computer software]. Wageningen, The Netherlands. https://www.noldus.com/observer-xt

Rushen, J., & de Passillé, A. M. (2012). The scientific assessment of the impact of housing on animal welfare: A critical review. *Canadian Journal of Animal Science*, 92(1), 1-18. https://doi.org/10.4141/cjas2011-100

Rushen, J., de Passillé, A. M., von Keyserlingk, M. A. G., & Weary, D. M. (2008). *The Welfare of Cattle*. Springer. https://doi.org/10.1007/978-1-4020-6558-3

Ventura, B. A., von Keyserlingk, M. A. G., Schuppli, C. A., & Weary, D. M. (2013). Views on contentious practices in dairy farming: The case of early cow-calf separation. *Journal of Dairy Science*, 96(9), 6105-6116. https://doi.org/10.3168/jds.2012-6040

Wolf, C. A., Tonsor, G. T., McKendree, M. G. S., Thomson, D. U., & Swanson, J. C. (2016). Public and farmer perceptions of dairy cattle welfare in the United States. *Journal of Dairy Science*, 99(7), 5892-5903. https://doi.org/10.3168/jds.2015-10619

Yeates, J. W., & Main, D. C. J. (2008). Assessment of positive welfare: A review. *The Veterinary Journal*, 175(3), 293-300. https://doi.org/10.1016/j.tvjl.2007.05.009

Ahloy-Dallaire, J., Espinosa, J., & Mason, G. (2018). Play and optimal welfare: Does play indicate the presence of positive affective states? *Behavioural Processes*, 156, 3-15. https://doi.org/10.1016/j.beproc.2017.11.011

Jensen, M. B., Vestergaard, K. S., & Krohn, C. C. (1998). Play behaviour in dairy calves kept in pens: The effect of social contact and space allowance. *Applied Animal Behaviour Science*, 56(2-4), 97-108. https://doi.org/10.1016/S0168-1591(97)00106-8

Bonneau, M., Vayssade, J. A., Troupe, W., & Arquet, R. (2020). Outdoor animal tracking combining neural network and time-lapse cameras. *Computers and Electronics in Agriculture*, 168, 105150. https://doi.org/10.1016/j.compag.2019.105150

Council Directive 2008/119/EC. (2008). Council Directive 2008/119/EC of 18 December 2008 laying down minimum standards for the protection of calves. *Official Journal of the European Union*, L 10, 7-13.

Domun, Y., Pedersen, L. J., White, D., Adeyemi, O., & Norton, T. (2019). Learning patterns from time-series data to discriminate predictions of tail-biting, fouling and diarrhoea in pigs. *Computers and Electronics in Agriculture*, 163, 104878. https://doi.org/10.1016/j.compag.2019.104878

EFSA AHAW Panel (EFSA Panel on Animal Health and Animal Welfare). (2023). Welfare of calves. *EFSA Journal*, 21(3), e07896. https://doi.org/10.2903/j.efsa.2023.7896

Færevik, G., Jensen, M. B., & Bøe, K. E. (2006). Dairy calves social preferences and the significance of a companion animal during separation from the group. *Applied Animal Behaviour Science*, 99(3-4), 205-221. https://doi.org/10.1016/j.applanim.2005.10.012

Jensen, M. B., Vestergaard, K. S., & Krohn, C. C. (1998). Play behaviour in dairy calves kept in pens: The effect of social contact and space allowance. *Applied Animal Behaviour Science*, 56(2-4), 97-108. https://doi.org/10.1016/S0168-1591(97)00106-8

Jensen, M. B., & Kyhn, R. (2000). Play behaviour in group-housed dairy calves: The effect of space allowance. *Applied Animal Behaviour Science*, 67(1-2), 35-46. https://doi.org/10.1016/S0168-1591(99)00113-6

Liu, L., Ni, J., Zhao, R., Shen, M., He, D., & Zhang, M. (2024). Design of a multiple object tracking algorithm for feeder pigs based on DeepSORT. *Computers and Electronics in Agriculture*, 217, 108576. https://doi.org/10.1016/j.compag.2023.108576

Marin, L. M., Neave, H. W., Nordgreen, J., & Jensen, M. B. (2024). Social housing facilitates behavioral synchrony in dairy calves. *Applied Animal Behaviour Science*, 270, 106091. https://doi.org/10.1016/j.applanim.2023.106091

Martin, P., & Bateson, P. (2007). *Measuring Behaviour: An Introductory Guide* (3rd ed.). Cambridge University Press. https://doi.org/10.1017/CBO9780511810893

Minderer, M., Gritsenko, A., Stone, A., Neumann, M., Weissenborn, D., Dosovitskiy, A., Mahendran, A., Arnab, A., Dehghani, M., Shen, Z., Wang, X., Zhai, X., Kipf, T., & Houlsby, N. (2023). Simple Open-Vocabulary Object Detection with Vision Transformers. *arXiv preprint arXiv:2205.06230*. https://doi.org/10.48550/arXiv.2205.06230

Psota, E. T., Pérez, L. C., Schmidt, T., Mote, B., & Velazco, J. I. (2020). Long-term tracking of group-housed livestock using keypoint detection and MAP estimation for individual animal identification. *Sensors*, 20(13), 3670. https://doi.org/10.3390/s20133670

Yang, H., Liu, E., Sun, J., Sharma, S., van Leerdam, M., Franceschini, S., Niu, P., & Hostens, M. (2025). A Computer Vision Pipeline for Individual-Level Behavior Analysis: Benchmarking on the Edinburgh Pig Dataset. *ArXiv Preprint ArXiv:2509.12047*. https://doi.org/https://doi.org/10.48550/arXiv.2509.12047

# Supplementary Material

## S1. Video Storage and Processing Technical Details

**Video File Formats:** Videos were recorded in H.264 codec format with resolution of 1920×1080 pixels (Full HD) at 25 frames per second. Each camera generated multiple video files per day, typically segmented into 1-hour clips to prevent file corruption in case of unexpected power interruption.

**File Organization Structure:** Upon transfer to external hard drives, files were organized hierarchically: `/FarmName/PenID/CameraID/YYYY-MM-DD/VideoFile.mp4`. This structure facilitated systematic processing and prevented file naming collisions across multiple farms.

**Storage Requirements:** Total raw video data from 16 farms over 17 hours per farm yielded approximately 3.8 TB of video footage. External hard drives (4 TB capacity) were used for primary storage, with backup copies maintained on university network storage.

# S2. Complete Ethogram with Operational Definitions

**Table S1. Complete ethogram of calf behaviours with operational definitions**

| Behaviour Category | Behaviour | Operational Definition | Source |
|---|---|---|---|
| **Locomotor Play** | Run | Rapid forward movement with all four legs off ground simultaneously during gait cycle | Jensen et al., 1998 |
| | Gallop | Fast running with characteristic three-beat gait pattern | Jensen et al., 1998 |
| | Buck | Rapid vertical leap with hindquarters elevated higher than forequarters, often with back arched | Jensen et al., 1998 |
| | Buck-kick | Buck combined with one or both hind legs extending backward | Jensen et al., 1998 |
| | Vertical leap | Jumping upward with all four legs leaving ground, without forward displacement | Jensen et al., 1998 |
| | Head-shake (play) | Vigorous side-to-side or up-and-down head movement occurring immediately before/after other play behaviours | Adapted from Jensen et al., 1998 |
| | Play-bounce | Bouncing movement with alternating vertical and forward motion, resembling pronking | Present study |
| **Social Play** | Frontal push | Two calves facing each other, pushing head-to-head or shoulder-to-shoulder with no aggressive intent | Jensen et al., 1998 |
| | Mounting | One calf placing front legs on another calf's back or rump | Present study |
| | Chase | One calf pursuing another calf at running speed, with pursuer maintaining visual contact with target | Present study |
| **Object Play** | Brush interaction | Rubbing, pushing, or scratching against stationary brush installed in pen | Present study |
| | Pen feature interaction | Investigating, pushing, or climbing on gates, feeders, or structural elements | Present study |
| **Straw Play** | Straw toss | Picking up straw with mouth and tossing head to throw straw into air | Present study |
| | Straw dig | Pawing at straw bedding with front hooves, displacing bedding material | Present study |
| **Non-Play States** | Management | External disturbance by human or non-calf animal entering camera view or pen | Gladden et al., 2020 |
| | Out of view | Calf partially or completely obscured from camera view for >0.5 seconds | Gladden et al., 2020 |
| | Individual | Calf lying alone, separated from group without social contact | Present study |
| | Milk feeding | Calf actively suckling from automated milk feeder | Present study |
| | Not Playing | All other behaviours not classified above (lying, standing, walking, feeding, ruminating, grooming) | Present study |

**Note:** Contextual interpretation was essential for distinguishing play from non-play expressions of similar movements (e.g., head-shaking for fly removal vs. play-related head-tosses).

## S3. Detailed Statistical Model Specifications

### *Base Model (Space Effect Only)*

The univariate linear mixed model examining space allowance effects on play duration was specified as:

**Model equation:**

$$\text{Play\_Duration}_{ij} = \beta_0 + \beta_1(\text{Space\_Group}_{ij}) + u_j + \varepsilon_{ij}$$

Where:

- Play_Duration$_{ij}$ = play duration (%OP) for calf i in farm j
- $\beta_0$ = overall intercept
- $\beta_1$ = fixed effect coefficient for space group
- Space_Group$_{ij}$ = categorical space allowance (seven 2 m² increment categories)
- $u_j$ = random intercept for farm j, capturing farm-level variation
- $\varepsilon_{ij}$ = residual error for calf i in farm j

**Estimation method:** Maximum likelihood estimation with variance components covariance structure

**Fixed effects testing:** Type III sums of squares

**Post hoc comparisons:** Least Significant Difference (LSD) tests for pairwise comparisons between space categories (21 comparisons among 7 categories)

### *Model Diagnostics*

**Normality assessment:** Standardized residuals were extracted and examined using:

- Histograms with overlay of normal distribution curve
- Q-Q plots comparing empirical quantiles to theoretical normal quantiles

- Shapiro-Wilk test (though with n=60, visual inspection prioritized over formal test)

**Homoscedasticity assessment:**

- Scatterplots of standardized residuals vs. predicted values
- Visual inspection for funnel patterns or systematic variance trends
- Levene's test for equality of error variances across space groups

(Interactions not retained due to non-significance and convergence issues)

### *Model Fit Statistics*

**Marginal R² (R²$_m$):** Proportion of variance explained by fixed effects only

```
R²m = (σ²total- σ²residual) / σ²total
```

**Conditional R² (R²$c$):** Proportion of variance explained by fixed + random effects

```
R²c = (σ²total - σ²residual) / σ²total (including farm variance)
```

**Intraclass Correlation Coefficient (ICC):**

```
ICC = σ²farm / (σ²farm + σ²residual)
```

Interpretation: Proportion of total variance attributable to farm-level clustering

**Significance threshold:** α = 0.05 for main effects; trends reported for 0.05 < p < 0.10

## S4. Metadata Integration and Data Engineering

The metadata integration pipeline merged manual behavioural annotations with computer vision outputs (frames, embeddings, tracking IDs) using Apache Spark for distributed processing. This section details the temporal alignment logic and final metadata structure.

## *Temporal Alignment Algorithm*

**Challenge:** Manual annotations used video time (seconds from start), while extracted frames used wall-clock timestamps. Accurate alignment required conversion between time formats.

**Solution:** Three-step conversion process:

1. **Extract video start timestamp** from filename (format: `FarmName_CameraID_YYYYMMDD_HHMMSS.mp4`)
2. **Convert annotation time** (seconds) to absolute timestamp:`absolute_time = video_start_time + annotation_seconds`

3. **Match to nearest frame** within 0.5-second tolerance window

**Spark implementation:**

```
# Load annotations with video start time
annotations_df = spark.read.csv("annotations.csv", header=True)
annotations_df = annotations_df.withColumn(
    "annotation_timestamp",
    col("video_start_time") + col("seconds_from_start")
)

# Load frame metadata
frames_df = spark.read.csv("frames_metadata.csv", header=True)

# Join on nearest timestamp (within 0.5s window)
matched_df = annotations_df.join(
    frames_df,
    (abs(col("annotation_timestamp") - col("frame_timestamp")) <= 0.5),
    "inner"
)
```

## *Label Hierarchy and Final Label Assignment*

Behavioural states were hierarchically organized:

1. **Primary level:** Management events (highest priority; override all other labels)
2. **Secondary level:** Play behaviours (Active Playing, Non-Active Playing)
3. **Tertiary level:** Out of view (lowest priority)

**Assignment logic:**

```
IF Management event occurs:
    Final_Label = "Management"
ELSE IF any Active Play behaviour present:
    Final_Label = "Active Playing"
ELSE IF Non-Active Play behaviour present:
    Final_Label = "Non-Active Playing"
ELSE IF Out of View:
    Final_Label = "Out of View" (excluded from training)
ELSE:
    Final_Label = "Not Playing"
```

This hierarchy ensured that important behavioural states (management disturbances, active play) were not overwritten by concurrent lower-priority states.

## *Final Metadata Structure*

The integrated CSV file contained the following columns:

| Column | Description | Data Type | Example |
|---|---|---|---|
| Timestamp | ISO 8601 timestamp with millisecond precision | String | 2024-06-15T08:23:45.120Z |
| Primary_Raw | Raw primary behaviour annotation | String | Gallop |

| Primary_Label | Classified primary behaviour | String | Active Playing |
|---|---|---|---|
| Secondary_Raw | Raw secondary behaviour (if simultaneous) | String | Buck |
| Secondary_Label | Classified secondary behaviour | String | Active Playing |
| ID | SAM2 tracking ID for calf | Integer | 3 |
| Final_Label | Hierarchically assigned label for training | String | Active Playing |
| Frame_Directory | Path to cropped frame image | String | /crops/FarmA/2024-06-15/frame_08234_id3.jpg |
| Embeddings_Directory | Path to DinoV2 embedding file | String | /embeddings/FarmA/2024-06-15/emb_08234_id3.npy |

**File size:** Approximately 150 MB per farm (CSV format), containing 200,000-300,000 rows per 17-hour observation period

**Quality assurance:**

- Automated checks verified no missing timestamps or frame paths
- Random sampling (5% of rows) manually inspected to confirm correct label assignment
- Cross-validation against original Observer XT files ensured no transcription errors

## S5. Classification Model: Architecture and Training Details

### *Model Architecture Overview*

The classification model employed a multi-layer perceptron (MLP) implemented in PyTorch, consisting of:

**Network Structure:**

- **Input layer:** 1024 dimensions (DinoV2 embedding size)
- **Hidden layer 1:** 1024 → 512 units, ReLU activation, 50% dropout
- **Hidden layer 2:** 512 → 256 units, ReLU activation, 50% dropout
- **Output layer:** 256 → 3 units (Active Playing, Non-Active Playing, Not Playing)
- **Total parameters:** 656,899 trainable parameters

**Architectural Rationale:** The two-hidden-layer architecture provided sufficient capacity to learn non-linear decision boundaries while remaining compact enough to train efficiently on the available dataset. ReLU activations introduced non-linearity, while dropout regularization (p=0.5) prevented overfitting by randomly deactivating neurons during training.

## *Training Configuration*

**Optimization:**

- **Optimizer:** Adam (Adaptive Moment Estimation)
- **Learning rate:** 0.001
- **Weight decay (L2 regularization):** $1 \times 10^{-5}$
- **Batch size:** 64 samples
- **Loss function:** Cross-entropy loss

**Cross-entropy loss** for sample with true class y and predicted logits **z**:

```
L = -log(exp(z_y) / Σᵢ exp(zᵢ))
```

This loss encourages high probability for the correct class while suppressing probabilities for incorrect classes.

**Data Loading:**

- Training data shuffled at each epoch to prevent order-dependent biases
- Validation/test data loaded without shuffling for consistent evaluation
- 4 parallel workers employed for batch prefetching

## *Regularization and Early Stopping*

**Three regularization mechanisms prevented overfitting:**

1. **Dropout (50%):** Randomly zeroed half of neuron activations during training, forcing the network to learn redundant representations and reducing co-adaptation between neurons.
2. **Weight Decay (L2 penalty):** Added regularization term proportional to squared weight magnitudes, encouraging simpler models with smaller parameters.
3. **Early Stopping:** Training monitored validation loss after each epoch. If validation loss failed to improve for 5 consecutive epochs (patience=5), training terminated and the best model checkpoint (lowest validation loss) was retained.

**Training Duration:** Maximum of 50 epochs permitted, though early stopping typically halted training earlier when validation performance plateaued.

## *Model Evaluation Procedure*

The best model (selected via early stopping on validation set) was evaluated on the held-out test set:

1. Model set to evaluation mode (dropout disabled)
2. Test embeddings passed through network to generate predictions
3. Predicted class determined by selecting class with highest output logit
4. Performance metrics computed using scikit-learn:
    a. Overall accuracy
    b. Per-class precision, recall, and F1-score
    c. Confusion matrix (both raw counts and normalized proportions)

## *Experiment Tracking and Reproducibility*

All training runs were tracked using **MLflow,** an open-source platform for managing machine learning experiments. MLflow automatically logged:

- **Hyperparameters:** Learning rate, batch size, dropout rate, architecture specifications

- **Metrics per epoch:** Training loss, training accuracy, validation loss, validation accuracy
- **Final test metrics:** Test accuracy, per-class precision/recall/F1-scores
- **Model artifacts:** Trained model weights, confusion matrices, metadata

This comprehensive logging enabled:

- Systematic comparison of different configurations
- Reproducible retrieval of trained models for evaluation and deployment
- Version control of model artifacts with environment specifications

**Note:** Complete implementation code, including detailed training loops and evaluation scripts, is available in the project's GitHub repository [https://github.com/Bovi-analytics/Individual-Behavior-Analysis-with-CV].